\documentclass[10pt,twocolumn,letterpaper]{article}

\usepackage{wacv}
\usepackage{times}
\usepackage{epsfig}
\usepackage{graphicx}
\usepackage{amsmath}
\usepackage{amssymb}
\usepackage{enumitem}
\usepackage{subcaption}
\usepackage{url}
\usepackage{multirow}
\usepackage{cleveref}
\usepackage{float}
\usepackage{soul}
\usepackage{subcaption}
\usepackage[font=small,labelfont=bf]{caption}
\usepackage{authblk}

\wacvfinalcopy 

\def\wacvPaperID{0040} 

\ifwacvfinal\fi
\begin{document}

\title{Beyond Spatial Auto-Regressive Models: Predicting Housing Prices with Satellite Imagery}

\author[1]{Archith J. Bency\thanks{archith@ece.ucsb.edu}}
\author[2]{Swati Rallapalli}
\author[2]{Raghu K. Ganti}
\author[2]{Mudhakar Srivatsa}
\author[1]{B. S. Manjunath}
\affil[1]{Department of Electrical and Computer Engineering, University of California, Santa Barbara, CA}
\affil[2]{IBM T.J. Watson Research Center, Yorktown Heights, NY}

\renewcommand\Authands{ and }

\maketitle
\ifwacvfinal\thispagestyle{empty}\fi

\begin{abstract}

 When modeling geo-spatial data, it is critical to capture spatial correlations for achieving high accuracy. Spatial Auto-Regression (SAR) is a common tool used to model such data, where the spatial contiguity matrix (W) encodes the spatial correlations. However, the efficacy of SAR is limited by two factors. First, it depends on the choice of contiguity matrix, which is typically not learnt from data, but instead, is assumed to be known apriori. Second, it assumes that the observations can be explained by linear models. 

In this paper, we propose a Convolutional Neural Network (CNN) framework to model geo-spatial data (specifically housing prices), to learn the spatial correlations automatically. We show that neighborhood information embedded in satellite imagery can be leveraged to achieve the desired spatial smoothing. An additional upside of our framework is the relaxation of linear assumption on the data. Specific challenges we tackle while implementing our framework include, (i) how much of the neighborhood is relevant while estimating housing prices?  (ii) what is the right approach to capture multiple resolutions of satellite imagery? and (iii) what other data-sources can help improve the estimation of spatial correlations? We demonstrate a marked improvement of 57\%  on top of the SAR baseline 
through the use of features from deep neural networks for the cities of London, Birmingham and Liverpool.


\end{abstract}

\section{Introduction}
\begin{figure}
\begin{center}
\includegraphics[width=\columnwidth]{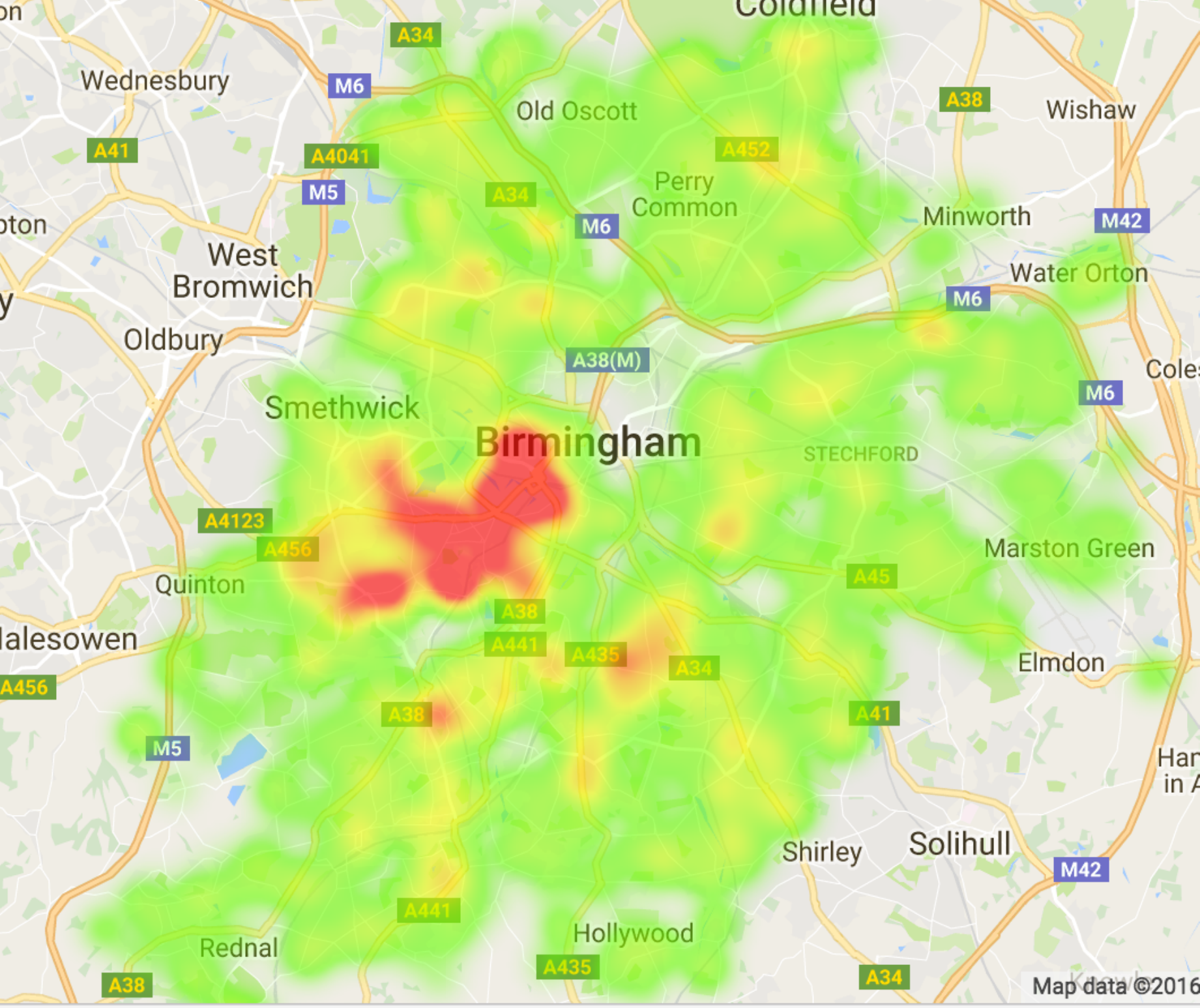}
\end{center}
   \caption{Housing sale price heat-map for the Birmingham, UK region. Red indicates high price areas and areas shaded blue are lower price areas. Spatial clusters of high property value areas and lower value areas can be seen. $\textit{Best viewed in color.}$ }
\label{fig:birminghamHeatMap}
\vspace{-0.1in}
\end{figure}

Housing Prices are important economic indicators of wealth and financial well-being in an urban scenario. In addition to house-specific metrics such as number of rooms and floors, square footage, and age,  the location of houses also have been shown to affect valuations\cite{kockelman1997effects, krupka2009neighborhood}. Neighbourhood effects include factors such as taxation policies, availability of transportation and general amenities. As a result, when we visualize housing prices on a geographical map, spatial clusters of high and low prices can be found as in Figure \ref{fig:birminghamHeatMap}, for the city of Birmingham, UK, where houses in adjacent regions might have similar prices. In addition to housing prices, other socially and economically relevant metrics such as crime-rates and pollution levels \cite{zwack2011modeling, steenbeek2011longitudinal} also demonstrate spatial clustering. Hence, models designed to represent such geo-spatial data need to capture the underlying spatial correlations. Traditionally, spatially dependent phenomena as those mentioned above are described using Spatial Auto-Regressive (SAR) models.

The SAR model combines neighbourhood relationships between samples and observed variables in a linear formulation to estimate spatially varying variables. The neighbourhood relationships are encoded in the form of a spatial contiguity matrix, and are often hand-designed with the help of domain expertise. The choice of the spatial contiguity matrix can lead to a trial and error process and leaves open the question of how to arrive at an optimal selection. 

In this paper, we present a mechanism to learn the neighborhood relationship patterns from the data. To do so, we find that incorporating features for house locations learnt from satellite images is very effective. In recent times, multiple commercial real-estate listings websites \cite{Zoopla, Zillow, Redfin, Trulia} store and display housing prices super-imposed on satellite imagery from mapping services \cite{BingMaps, GoogleMaps}. The existence of such data makes large quantities of satellite images available with associated house prices. Satellite images provide a `bird's eye view' of a location and the neighbourhood it is situated in. In addition to the top-down appearance of a house, they also provide contextual information about the immediate and larger area of surroundings. We train Deep Convolutional Neural Networks (DCNNs) to discriminate between images learnt at different spatial scales corresponding to more and less affluent locations in a given city. The features learnt in the process are combined with house specific attributes through an estimator to arrive at a price estimate. The main contributions of this paper are:
\begin{itemize}[leftmargin=*]
    \item{We present a method where neighbourhood information for geo-spatial samples is learnt implicitly through satellite image features }
    \item{We examine the impact of using neighbourhood information at multiple geo-spatial scales on housing prices estimation }
\end{itemize}

The remainder of this paper is organized as follows. We present the related work in Section \ref{section:relatedWorks}. Our approach is described in Section \ref{section:main} with a brief review of the SAR model, deep feature extraction, multi-modal fusion and price estimation. In the subsequent Section \ref{section:experiments}, we characterize experiments conducted for price estimation and describe the data-sets, metrics and results. We discuss the results and conclude the paper in Section \ref{section:discussions}.


\section{Related Work}\label{section:relatedWorks}
We overview the related works by broadly diving them into the following categories.
\paragraph*{Housing Price Estimation:} is a classical problem in the field of spatial econometrics \cite{bourassa2007spatial, osland2010application, lesage2009introduction}. These methods utilize attributes such as house square-footage, number of rooms, number of floors, age of the house, garage space etc. Extracting such detailed information for large data-sets would be a tedious task. Moreover, the spatial dependence of samples on each other is modelled using the SAR model. The choice of spatial contiguity matrix (W) used by the SAR model is \textit{hand designed} (as opposed to learning from the data) either using Delaunay Triangulation, k-Nearest Neighbours computation or Quasi-local correlation functions \cite{lesage2009introduction, chen2012four}. While the work in\cite{chopra2007discovering} learns both sample level and spatially smooth manifold features from housing price data, it incorporates only non-visual features and needs fine-grained data such as type of heating and type of air conditioning amongst others which might be difficult to obtain on a large-scale for urban areas.

\paragraph*{Applications of Street View Imagery:} There has been increased attention in the computer vision community on the problem of urban scene analysis. Features learnt using Deep Convolutional Neural Networks \cite{krizhevsky2012imagenet, simonyan2014very} have been shown to be effective at representing complex contextual information by learning from large-scale data-sets. \cite{CVPR14_Khosla, ordonez2014learning, arietta2014city, bessinger2016quantifying} investigated the correlation between visual features extracted from the street view imagery of cities and the high-level human perceptions on safety, wealth, directions to ubiquitous city landmarks and housing prices. These works have focused on utilizing street view images, which  provide rich visual information in the immediate vicinity of houses, but do not describe a larger neighbourhood which is the case for satellite images.

\paragraph*{Applications of Satellite Imagery:} Satellite images have been analysed in the context of road detection \cite{mnih2010learning, hu2007road}, predicting poverty \cite{jean2016poverty}, object detection \cite{cheng2016survey} and tracking \cite{meng2012object}. In this work, we utilize  images from this modality for the problem of housing price prediction. We further study the impact of using satellite images from different zoom levels on the accuracy of our models.


\section{Estimating Housing Prices}\label{section:main}

\begin{figure*}
\begin{center}
\includegraphics[width=2\columnwidth]{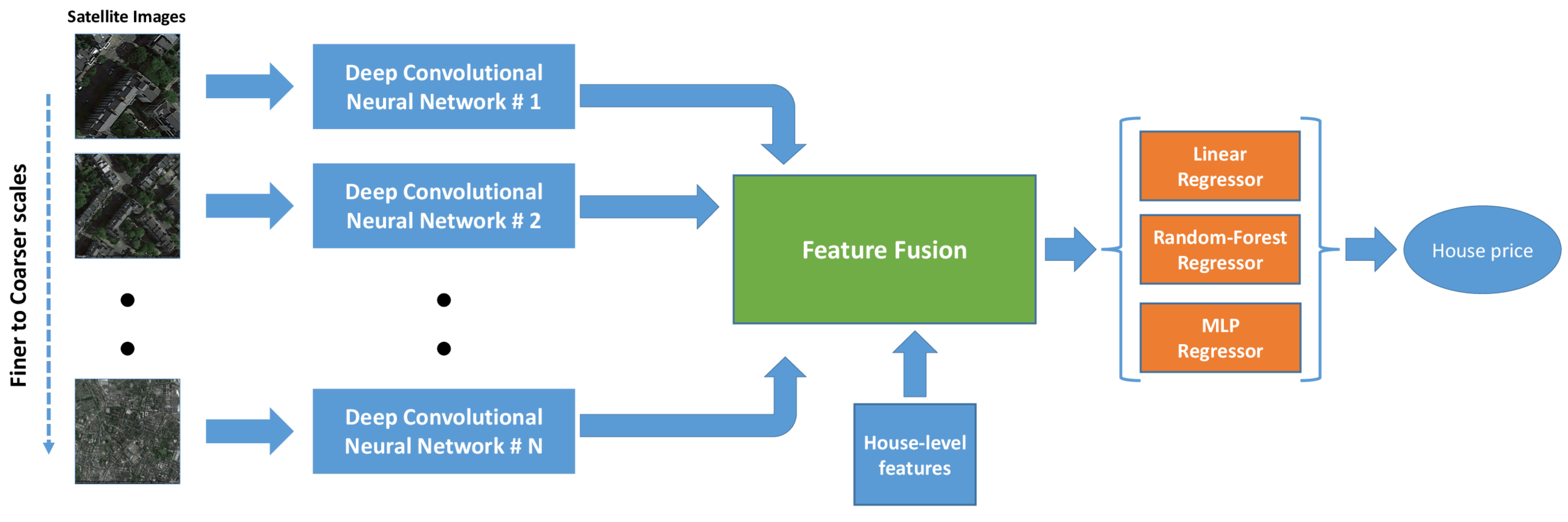}
\end{center}
\vspace{-0.1in}
   \caption{Satellite images of regions around test house samples are extracted from finer to coarser scales. Deep CNN features are extracted to get neighbourhood information and are fused with house-level explanatory variables through concatenation. The joint description  of a house and it's neighbourhood is used to estimate it's price through regressors. $\textit{Best viewed in color.}$ }
\label{fig:block_diagram}
\vspace{-0.1in}
\end{figure*}

\subsection{Background: SAR model}
SAR models are the traditional methods used to describe geo-spatial data. Dependent variables (to be estimated) are modelled as a weighted sum of dependent variable values of geo-spatial neighbours and the sample's observed variables. Mathematically, the SAR model is represented as:
\begin{equation}
\vspace{-0.1in}
\begin{split}
y & = \rho Wy + X \beta + \epsilon \\
\epsilon & \sim N(0, \sigma^2 I_{n})
\end{split}
\label{eq:SAR}
\vspace{0.2in}
\end{equation}

\noindent $y$ denotes the dependent variable of size $n \times 1$, $X$ of size $n \times k$ represents the observed variables, $W$ is the $n \times n$ row normalized spatial contiguity matrix, $\rho$ is the coefficient of spatial dependence for $y$, $\beta$ of size $k \times 1$ signifies the influence of observed variables and  $\epsilon$ is the error term modelled as a zero mean Gaussian distribution.

The parameters of the model, $\rho$ and $\beta$ are learnt through Maximum Likelihood estimation \cite{lesage2009introduction}. The choice of $W$ defines how neighbouring samples influence each other. $W$ is constructed as a sparse matrix where $W_{i,j} = 1$ for samples $i$ and $j$ which are neighbours. One criterion for samples  to be neighbours is when they are within a distance of $r_W$ of each other \cite{chen2012four}. Another method of neighbourhood definition has  been designed through Delaunay Triangulation \cite{lesage2009introduction}. Two samples which share an edge of a constructed triangulation are considered to be neighbours. It is apparent that the choice of $W$ is highly dependent on domain expertise.

Under the condition of $|| \rho W || < 1$, equation \ref{eq:SAR} is  re-written as a power series expansion:
\begin{equation}
y = \sum_{i = 0}^{\infty} \rho^{i} W^{i} (X\beta + \epsilon)
\label{eq:SARexpansion}
\end{equation}

\noindent The equation can be interpreted as a decomposition of $y$ in terms of increasing powers of $W$. Since $W$ denotes spatial contiguity, terms with higher powers of $W$ represent contribution of sample's larger neighbourhoods in the value of $y$. In this work, we aim to emulate the effect of larger spatial neighbourhoods on dependent variables such as house prices through satellite images covering progressively larger geo-spatial areas, which provide a implicit and rich modality of information, instead of a hand-designed choice of $W$.


\subsection{System Architecture}
System architecture of the proposed method for estimating housing prices is presented in Figure \ref{fig:block_diagram}. We detail the various components of the system below.

\paragraph*{Data Sources:} Our framework leverages multiple modalities of data. (i) \textit{House Attributes:} First, we construct a database of house samples with latitude, longitude and house attributes from publicly available sources \cite{Zoopla, Zillow}. House attributes are composed of number of bedrooms, bathrooms, reception rooms and floors. (ii) \textit{Satellite Imagery:} Second, using the latitude and longitude coordinates, we query satellite images centered around the coordinate value \cite{GoogleMaps, BingMaps}. The images are sampled at different geo-spatial resolutions, keeping the image sizes at a constant value. The finer resolution scales result in images spanning the extent of individual houses, whereas coarser resolution images can span several city blocks or city districts.

\begin{figure}[t]
\centering
\includegraphics[width=\columnwidth]{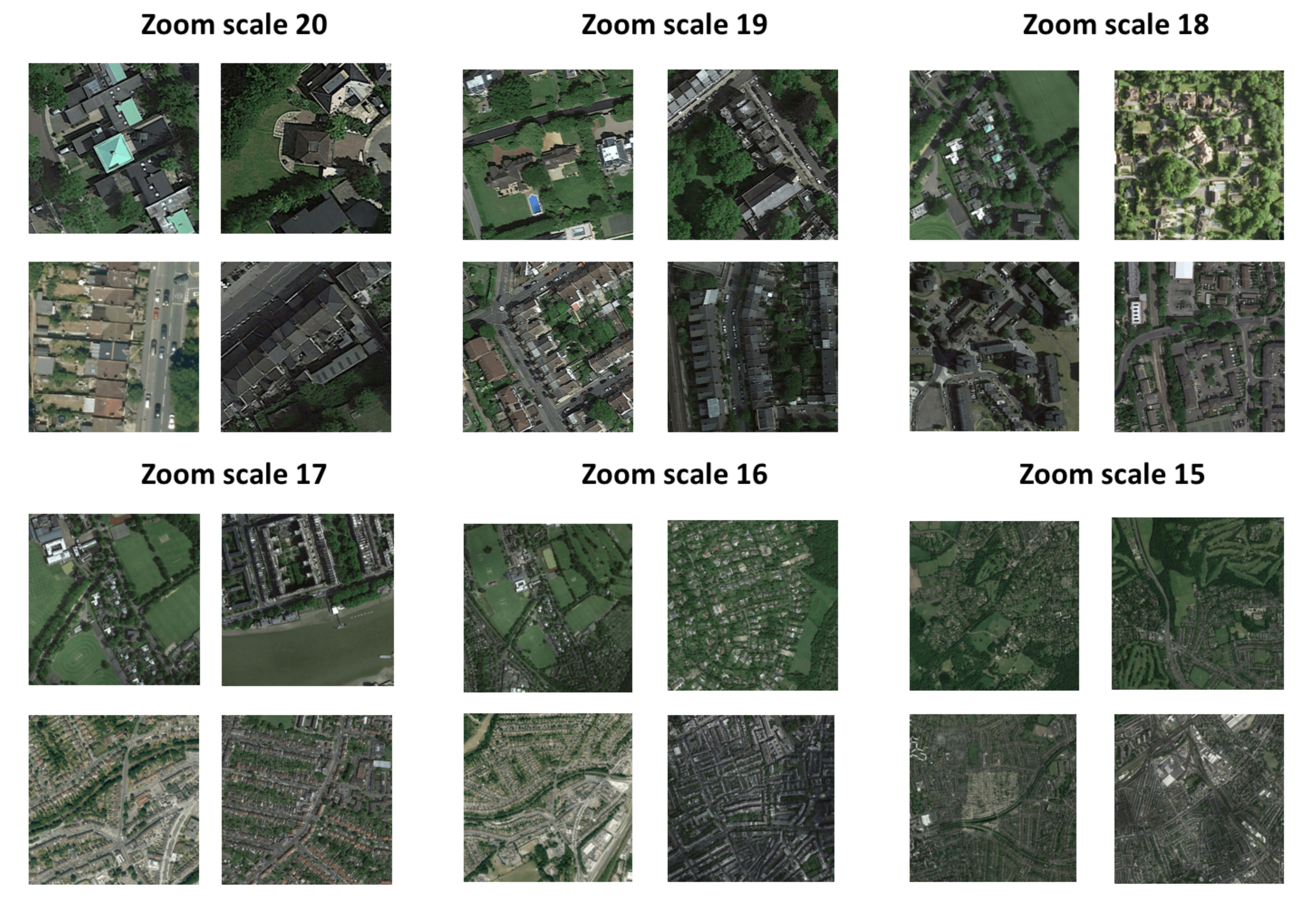}
   \caption{Examples of satellite images from the London city sale data-set. For each zoom value, the top row is constituted by examples from the top 10$\%$ of the data-set in terms of house price and the bottom row contains examples from the bottom 10$\%$. $\textit{Best viewed in color.}$ }
\label{fig:satelliteImageExamples}
\end{figure}

\paragraph*{Feature Extraction From Satellite Imagery:} 
Our next goal is to be able to extract features from the satellite imagery to capture the neighborhood effects. To this end, we leverage Deep Convolutional Neural Networks (DCNNs). However, through our experiments, we found that training DCNNs directly for predicting housing prices is challenging. First, the models often converge slowly and overfit for the training data. This is likely due to the noise in the house prices and the aerial imagery not being able to distinguish between houses with slight differences in their asking prices. Therefore, to learn features that can generalize well to other data-sets, we use \textit{transfer learning}  and train the feature extraction pipeline for a similar but simpler problem. Through our experiments, we found that the binary classification problem of distinguishing between expensive and cheap houses, learns features that can generalize with excellent accuracy to other data-sets and other related tasks. The two classes we use for this task, are the top $\delta\%$ and the bottom $\delta\%$ of the training data-set in terms of price. Intuitively, expensive houses within a given city tend to exist in neighborhoods with larger backyard and green-space and water bodies such as ponds and swimming pools, whereas cheaper houses tend to be located in compact neighborhoods where the houses are adjacent to each other with concrete and roads occupying more space. These differences are apparent to the human eye in satellite images and we designed the choice of class definitions with this factor in mind. This choice of design for the classes is intended to enable networks to learn features which are sensitive to price variations. Some examples of satellite images used for training are provided in Figure \ref{fig:satelliteImageExamples}. 

Second, learning and combining features from different zoom levels of satellite imagery is non-trivial. One straightforward approach is to use a single network that can process all the zoom levels. However, our experiments showed that this mixes up the features, rendering poor accuracy. Therefore, as shown in Figure \ref{fig:block_diagram}, we tackle this challenge by training a separate DCNN to learn features from each of the zoom levels. Further details on how the DCNNs were trained are provided in Section \ref{sec:DCNNTraining}.


\paragraph*{House Price Estimation:} The features extracted from the deep networks are concatenated with the house attributes into a feature vector $x_{feat}$. The resultant vector is then used to regress on housing prices through an estimator. In our experiments, we train multiple models, namely, (i) Linear, (ii) Random Forest and (iii) Multi-layer Perceptron regressors to understand and compare their effectiveness in estimating the housing prices. 

\begin{figure*}
\centering
\begin{subfigure}{.35\textwidth}
 
  \includegraphics[width=\textwidth]{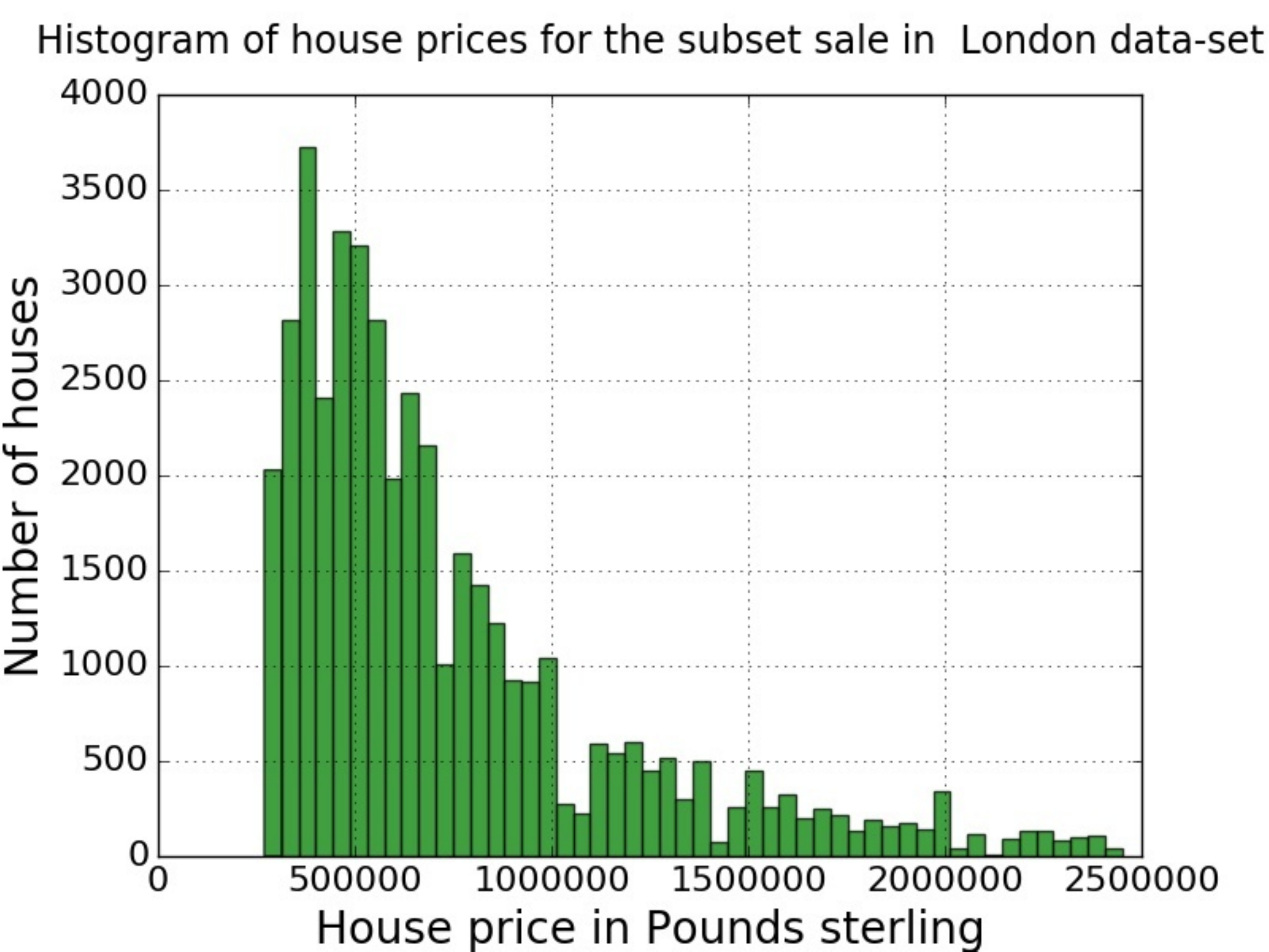}
  \caption{Sale sub-set}
  \label{fig:londonSale}
\end{subfigure}%
\begin{subfigure}{.35\textwidth}
 
  \includegraphics[width=\textwidth]{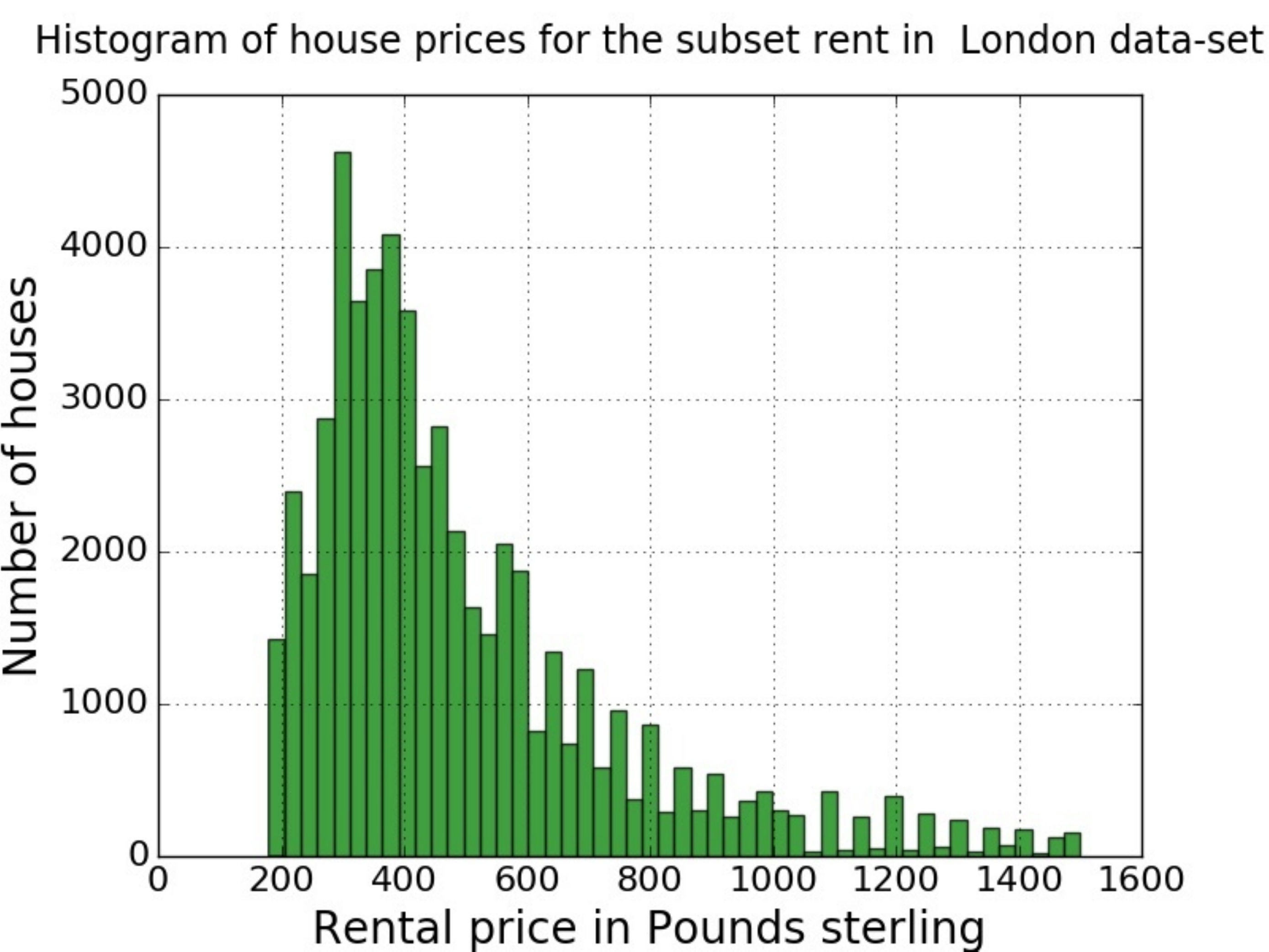}
  \caption{Rent sub-set}
  \label{fig:londonRent}
\end{subfigure}

\caption{Distribution of house price and rental values for the city of London in our collected data-set.}
\label{fig:londonPriceDistPlots}
\end{figure*}

\subsection{Point of Interest Data}
In order to validate the hypothesis that satellite images provide information regarding neighbourhoods for the task of price estimation, we also consider point of interest data. A point of interest is a location on a map which has economic, social or cultural value. Examples include fire-stations, restaurants, shopping centers, places of worship and bus stops. Each point of interest $x$ is hence described by  latitude and longitude coordinates and a place type $f_x$. The set of all place types is denoted by $P$. In order to describe a house location $h$ in terms of  point of interest information, we generate a feature vector $p_{h, r}$ which represents the number of instances of place type within a distance $r$. Mathematically,
\begin{equation}
\begin{split}
p_{h,r} &= [p_{h,r,t}] \ \ \forall \ \ t \in P \\
p_{h,r,t} &= \{|x| \mid d_{h,x} < r \ , \ f_x = t\}
\end{split}
\label{eq:POI}
\end{equation}
\noindent where $d_{h,x}$ is the haversine distance between locations $h$ and $x$ and $|A|$ represents the cardinality of a set $A$. In our experiments, we utilize $p_{h,r}$ as a feature vector describing the neighbourhood of the sample $h$. The parameter $r$ serves as a hyper-parameter, for which the optimum value is deduced using cross-validation.


\section{Experiments}\label{section:experiments}
\subsection{Data-sets}
\paragraph*{House price data-set:}  We extract housing price listings from a web-based source. Each listing entry consists of house-level details such as house sale price or rental value in Pounds sterling (\pounds), latitude, longitude, number of bathrooms, number of bedrooms, number of floors, number of reception rooms, listing status, street address, and a textual description. We extract the first eight attributes for each entry. The listing status for an entry can have two values, `rent' or `sale'. Since the price ranges for the sale and rent subsets do not overlap, we conduct our experiments separately on each subset. For each of the data-sets, 10\% of the samples are randomly picked without substitution and are designated as the \textit{test} set and the rest are marked as the \textit{training} set. 

We use data from the cities of London, Birmingham and Liverpool. The details regarding the number of samples in data-sets are presented in Table \ref{table:data-setSize}. We filter out samples with the top and bottom 2 $\%$ of the data-sets in terms of house sale price or rental value to prune out spurious or outlier entries. Due to relatively small number of samples in the `rent' subset for Birmingham and Liverpool, we do not analyse rental price estimations for these cities. Distribution of prices in the London city data-set are presented in Figure \ref{fig:londonPriceDistPlots}, with  price statistics for all data-sets shown in Table \ref{table:data-setRange}.

\paragraph*{Satellite Images:} Using the house coordinates, we extract RGB satellite images through the Google Maps \cite{GoogleMaps} web service. Google Maps allows queries in the format of (latitude, longitude, zoom value, image size). Latitude and Longitude specify the center location of the satellite image, zoom values are integer values which specify the geo-spatial resolution of retrived images. In this experiment, we specify queries with zoom values of 15, 16, 17, 18, 19 and 20, and the image size is fixed at 600$\times$600.  The images cover areas of 3.175, 0.794, 0.199, 0.048, 0.012 and 0.003
sq. kms. respectively. Considering the much larger average radius of the Earth, which is approximately 6371.0 km, we ignore  effects of Earth's curvature while calculating extents.

\paragraph*{Point of Interest Data:} In addition to satellite images, we have extracted place of interest data from the Google Places \cite{GooglePlaces} web service. The Places service provides an interface to query places and business of interest described by tags within a specified radius from a latitude-longitude pair. Tags act as classes of places that are desired, with the service offering 86 pre-set tags.\footnote{The complete list of tags supported by the Google Places service is available at \url{https://developers.google.com/places/supported_types}.}  Examples of tags include \textit{cafe}, \textit{beauty\_salon}, \textit{clothing\_store} and \textit{post\_office}. 
For each tag, our queries cover the entire area of the cities in our data-set.
The total number of places of interest retrieved for each city is listed in Table \ref{table:data-setSize}. We construct a ball tree for each tag to facilitate efficient nearest-neighbour radial searches \cite{kibriya2007empirical}.

\begin{table}[]
\vspace{-0.1in}
\footnotesize
\centering
\begin{tabular}{|c|c|c|c|}
\hline
\textbf{City} & \textbf{Sale} & \textbf{Rent} & \textbf{Place of Interest} \\ \hline
London        & 43037         & 55700         &     227800\\ \hline
Birmingham    & 3212          & -          &     50306\\ \hline
Liverpool     & 5004          & -          &      32878 \\ \hline
\end{tabular}
\caption{Number of sale and rent listings, and place of interest entries in city data-sets}
\label{table:data-setSize}
\vspace{-0.1in}
\end{table}

\begin{table}[]
\footnotesize
\centering
\begin{tabular}{|c|c|c|l|c|c|l|}
\hline
\multirow{2}{*}{\textbf{City}} & \multicolumn{3}{c|}{\textbf{Sale}} & \multicolumn{3}{c|}{\textbf{Rent}} \\ \cline{2-7} 
 & \begin{tabular}[c]{@{}c@{}}Mean \\ price \\ (\pounds)\end{tabular} & \multicolumn{2}{c|}{\begin{tabular}[c]{@{}c@{}}Median \\ price \\ (\pounds)\end{tabular}} & \begin{tabular}[c]{@{}c@{}}Mean\\  price\\  (\pounds)\end{tabular} & \multicolumn{2}{c|}{\begin{tabular}[c]{@{}c@{}}Median\\ price \\ (\pounds)\end{tabular}} \\ \hline
London & 743217.10 & \multicolumn{2}{c|}{599997.04} & 492.02 & \multicolumn{2}{c|}{415.00} \\ \hline
Birmingham & 181958.31 & \multicolumn{2}{c|}{154999.94} & - & \multicolumn{2}{c|}{-} \\ \hline
Liverpool & 151829.73 & \multicolumn{2}{c|}{133252.35} & - & \multicolumn{2}{c|}{-} \\ \hline
\end{tabular}
\caption{Statistics of sale and rent prices in city data-sets}

\label{table:data-setRange}
\vspace{-0.2in}
\end{table}

\subsection{Deep Convolutional Neural Network training}\label{sec:DCNNTraining}
In order to engineer features from the satellite images data-set, we train DCNNs to classify samples into the top and bottom $\delta\%$ in terms of sale price in case of the `sale' subset and rental price in case of the `rent' subset as explained in Section~\ref{section:main}. The value of $\delta$ is set to 10 in our experiments to strike a balance between requirements of keeping  reasonable number of samples available for training the DCNNs, and of keeping the image classes visually distinguishable. The choice of classes is also designed such that the features are sensitive to visual cues which distinguish between expensive and cheap houses and properties. Due to the limited size of our housing price data-sets, we fine-tune the Inception v3 \cite{szegedy2015rethinking} DCNN, which is a state-of-the-art image classifier trained on the larger Imagenet data-set \cite{ILSVRC15}. Since both data-sets consist of natural images, the generic convolutional features learnt in the early layers are re-used  and the final layers which are more task-specific are re-learnt for the new task \cite{fineTuning}.

The final blocks of convolutional filters and fully connected layers (\textit{mixed\_8$\times$8$\times$2048b}, \textit{logits}) are re-trained, with rest of layers kept fixed to the values learnt from Imagenet. The fully connected layer \textit{logits} is modified to generate features of dimension 256 which act as input to the final logistic classifier.

Considering the number of samples in our city data-sets from Table \ref{table:data-setSize}, we train DCNNs for the London data-set alone and conduct experiments on the efficacy of applying them on Liverpool and Birmingham in Section \ref{subsection:results}.
We train six neural networks, one for each zoom value and keep a fixed learning rate of 0.001. The \textit{train} subset is split in a 90:10 ratio into the classification \textit{train} and \textit{test} subsets. 
Table~\ref{table:classification} shows that the classification accuracy of these networks, is above 90\% for houses on sale and between 83\% and 89\% for houses put up for rent. Note that in this case, the chance performance is 50\%. \textit{The classification accuracy indicates that the networks are able to learn features that can distinguish between expensive and cheaper houses based on satellite images at each zoom level.}

For feature-extraction, we remove the logistic classifier layer from the neural network and the 256 dimensional features are used for price estimation. DCNN models were trained using Tensorflow \cite{abadi2016tensorflow} on a server configured with a Xeon E5-2630 CPU and a single NVIDIA Titan X GPU.

\begin{table*}[t]
\footnotesize
\centering
\begin{tabular}{|c|c|c|c|c|c|c|}
\hline
\textbf{Zoom value} & \textbf{15} & \textbf{16} & \textbf{17} & \textbf{18} & \textbf{19} & \textbf{20} \\ \hline
\textbf{Classification accuracy : Rent subset} & 88.96\% & 87.34\% & 86.04\% & 85.52\% & 84.32\% & 83.59\% \\ \hline
\textbf{Classification accuracy : Sale subset} & 90.36\% & 90.30\% & 90.49\% & 90.10\% & 90.30\% & 90.69\% \\ \hline
\end{tabular}
\caption{ Performance for classifying between the top and bottom 10 \% of samples in terms of house prices for different zoom values in the London data-set.}

\label{table:classification}
\vspace{-0.2in}
\end{table*}

\subsection{Estimators}
We use three different types of estimators on features to regress on housing prices. The estimators used are (i) Linear, (ii) Random Forest (RF) and (iii) Multi-layer Perceptron (MLP) regressors. For RF regressors, we use 40 decision tree estimators. In the case of MLP regressors, we use 2 hidden layers with the number of nodes set as (500,100) for the London data-set and (50,10) for Liverpool and Birmingham data-sets. These choices of hyper parameters were arrived at by minimizing root mean square error on a random 90:10 train and validation split on the \textit{train} subset. In order to account for stochasticity in training of RF and MLP regressors, we train 10 instances of the estimators and present mean and standard deviation for the result metrics in Section \ref{subsection:results}. All estimators were trained and tested using the Scikit-learn~\cite{scikit-learn} python library.

\subsection{Metrics}

We use two standard regression metrics to report the efficacy of our proposed method in estimating house sale or rental values: (i) Root Mean Square Error (RMSE) and (ii) Coefficient of Determination, also known as $R^{2}$. $R^{2}$ measures how well the variation in data is explained by the regression model. If we consider $y$ to be a vector of true price values and $Y$ to be the estimated values,
\begin{equation}
\vspace{-0.1in}
R^2 \ = \ 1 - \frac{E[(Y-y)^2]}{E[(y-\bar{y})^2]}
\end{equation}
\noindent with $\bar{y}$ representing the mean value of $y$. $R^2$ can take on values in the range $(-\infty, 1.0]$, with 1.0 indicating that the model is able to perform a perfect fit on test data.

\subsection{Results}\label{subsection:results}

\begin{figure}[htp]

\begin{minipage}{\columnwidth}
  \centering
  \includegraphics[width=\columnwidth]{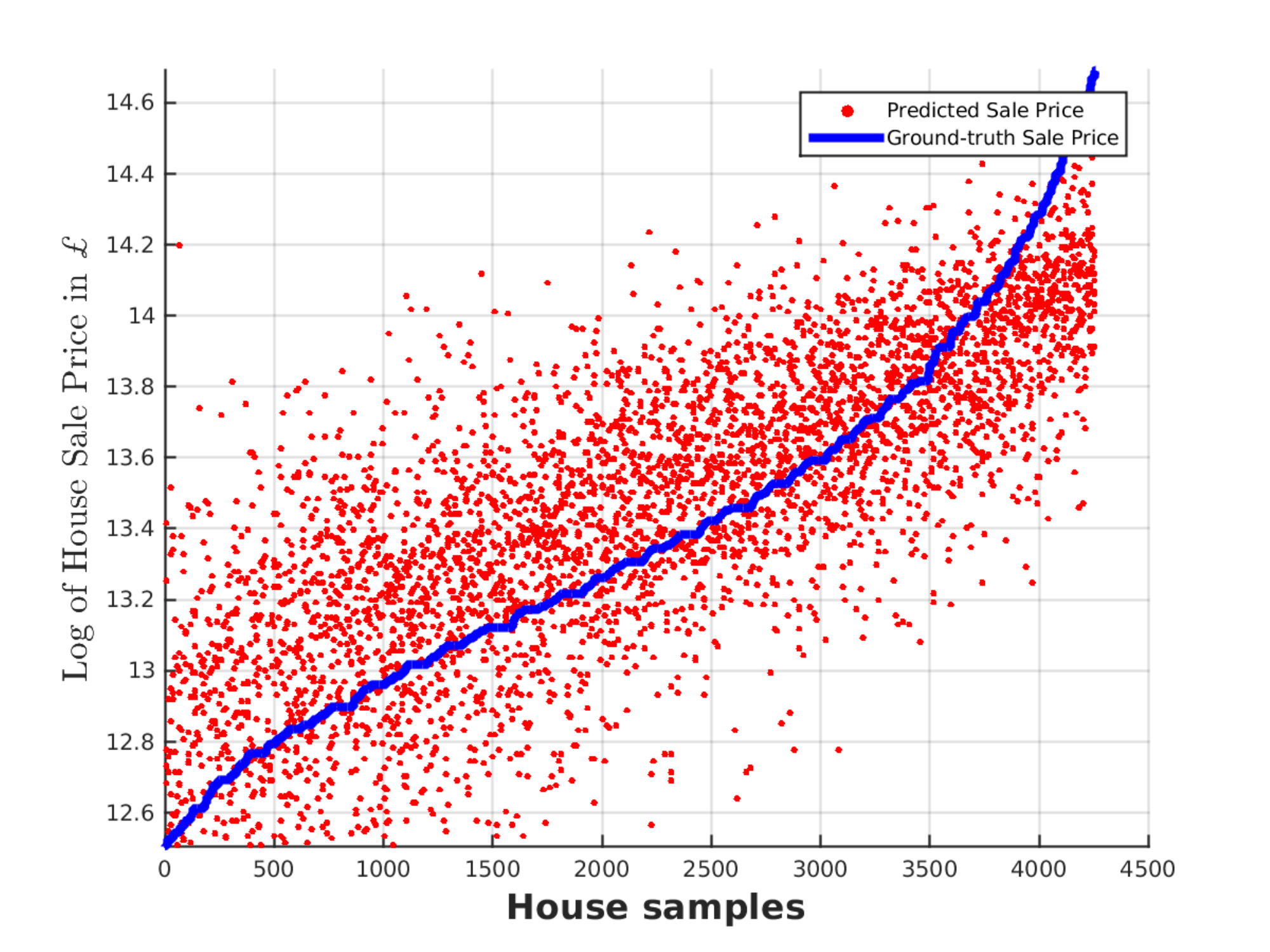}

  \vfill
  \includegraphics[width=\columnwidth]{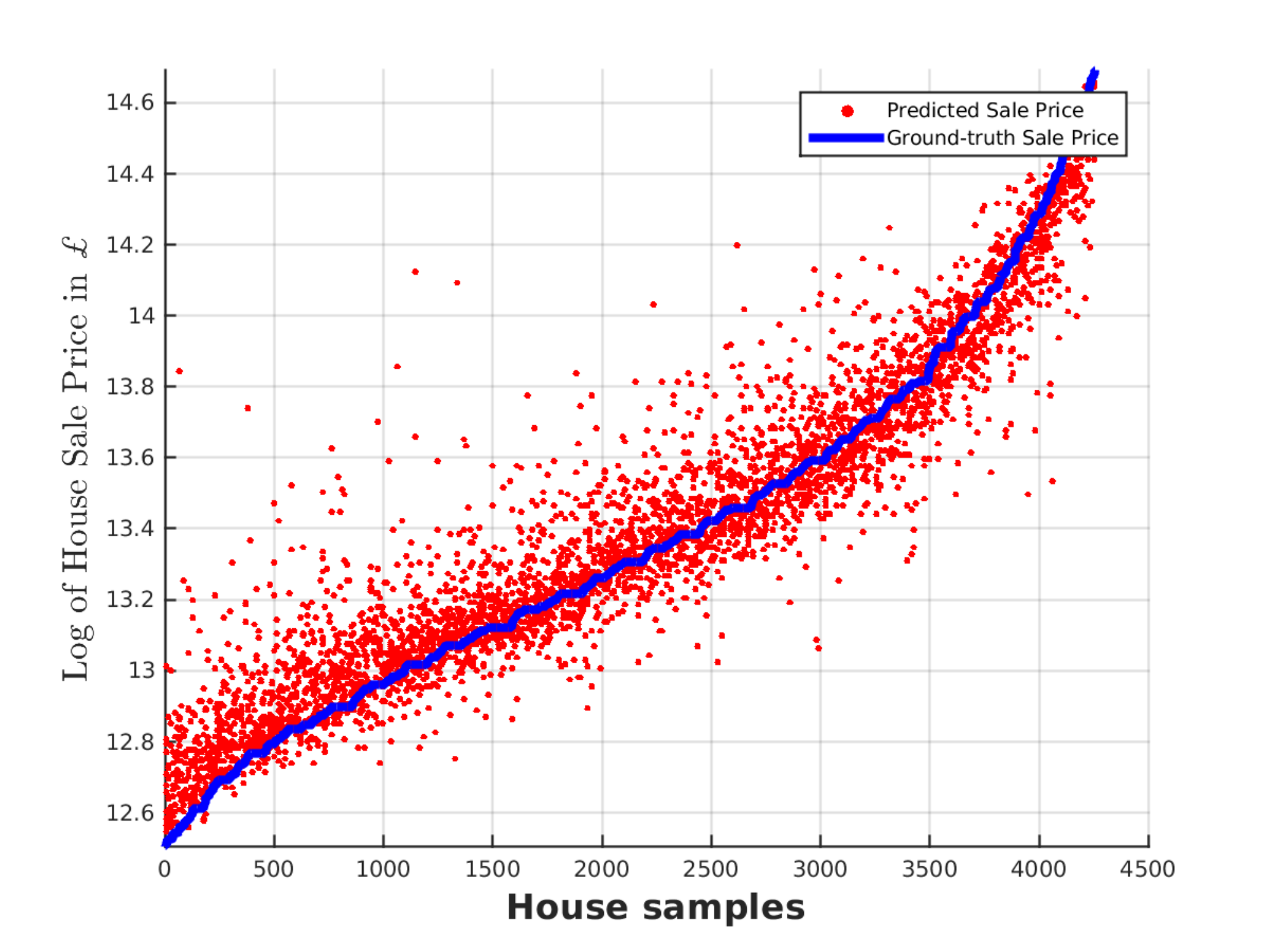}
\end{minipage}

\caption{Top: Logarithm of SAR model predicted sale housing prices for the
London city data-set is shown as a scatter plot. Bottom: Logarithm of proposed method predicted sale housing prices for the
same data-set with House Attribute and Deep Features using Random Forest regression .}
\label{fig:SARvsPropMethod}
\end{figure}

\begin{table}[!t]
\scriptsize
\centering
\begin{tabular}{|r|c|c|}
\hline
\multicolumn{1}{|c|}{\multirow{2}{*}{\textbf{\begin{tabular}[c]{@{}c@{}}Deep Feature \\ Zoom Values\end{tabular}}}} & \multicolumn{2}{c|}{\textbf{London}} \\ \cline{2-3} 
\multicolumn{1}{|c|}{} & \textbf{Rent} & \textbf{Sale} \\ \hline
\multicolumn{3}{|c|}{\textbf{RMSE ($\pounds$)}} \\ \hline
HA + DF (20) & 81.20 $\pm$ 0.71 & 156821.14 $\pm$ 1534.30 \\ \hline
HA + DF (19, 20) & 80.65 $\pm$ 0.47 & 157200.04 $\pm$ 1617.62 \\ \hline
HA + DF (18, 19, 20) & 79.14 $\pm$ 0.63 & 155761.95 $\pm$ 1512.96 \\ \hline
HA + DF (17, 18, 19, 20) & \textbf{79.34 $\pm$ 0.54} & 156010.72 $\pm$ 1209.60 \\ \hline
HA + DF (16, 17, 18, 19, 20) & 79.59 $\pm$ 0.24 & 144524.24 $\pm$ 1050.83 \\ \hline
HA + DF (15, 16, 7, 18, 19, 20) & 79.78 $\pm$ 0.63 & \textbf{127303.06 $\pm$ 1634.04} \\ \hline
\multicolumn{3}{|c|}{\textbf{$R^2$}} \\ \hline
HA + DF (20) & 0.8905 $\pm$ 0.0019 & 0.8647 $\pm$ 0.0026 \\ \hline
HA + DF (19, 20) & 0.8925 $\pm$ 0.0012 & 0.8640 $\pm$ 0.0028 \\ \hline
HA + DF (18, 19, 20) & 0.8960 $\pm$ 0.0016 & 0.8665 $\pm$ 0.0026 \\ \hline
HA + DF (17, 18, 19, 20) & \textbf{0.8955 $\pm$ 0.0014} & 0.8661 $\pm$ 0.0021 \\ \hline
HA + DF (16, 17, 18, 19, 20) & 0.8948 $\pm$ 0.0006 & 0.8851 $\pm$ 0.0016 \\ \hline
HA + DF (15, 16, 7, 18, 19, 20) & 0.8946 $\pm$ 0.0008 & \textbf{0.9108 $\pm$ 0.0022} \\ \hline
\end{tabular}
\caption{RMSE and $R^2$ for different neighbourhood zoom values with Random Forest regression. Lower values for RMSE and higher values for $R^2$ indicated superior performance with the method with best results highlighted in bold. HA and DF stand for House Attributes and Deep Features respectively. }
\label{table:neighbourhoodSize}
\end{table}


\paragraph*{Comparison with SAR and other base-lines:} We compare  performance of the proposed method with the SAR technique in Table \ref{table:SARvsPropRMSE&R2}. The observed variables matrix for SAR, $X$ from equation \ref{eq:SAR}, consists of the house attributes and point of interest features with $r$ set to 2.0 km. We utilize an implementation of SAR from \cite{lesage2009introduction}. The choice of $W$ is derived through Delaunay Triangulation (DT) and K-Nearest Neighbour (KNN). For the proposed method, concatenation of house attributes and deep features from satellite images from all zoom values is used to arrive at the results though Linear, Random Forest and MLP estimators. Even though the proposed method with Linear Regression and SAR are both linear models, we arrive at superior results through the usage of neighbourhood information in the form of satellite image features. Random Forest and MLP regressors further improve upon the estimation performance. A visualization of the comparison between predictions from the SAR model and the proposed method is presented in Figure \ref{fig:SARvsPropMethod}. 

In Table~\ref{table:ICIP16Comp}, we compare our approach with \cite{bessinger2016quantifying}, a recent method which estimates housing prices through house attributes (HA) and features extracted from street-level images through DCNNs. We use the data-set released by authors of the above work. Real estate data for houses in  Fayette County, Kentucky, USA are provided with details regarding house location and attributes such as \textit{Tax rates}, \textit{Acres} and \textit{Total Rooms} along with a train-test split. As the networks used for feature extraction in \cite{bessinger2016quantifying} are fine-tuned versions of VGG-16 architecture \cite{simonyan2014very}, we also utilize the same architecture with training procedure as described in Section \ref{sec:DCNNTraining}.   We demonstrate superior performance through a  13.5\% reduction in RMSE through using HA + IF features. In the case of using Image Features, which in our case are the features extracted through DCNNs on the satellite images extracted at different zoom values, we achieve a larger reduction of 34.5\% in RMSE. The results show that features through multi-scale satellite imagery are able to better explain housing price variations. Random Forest regressors with 720 estimator trees are used in this result.

\begin{table}[]
\footnotesize
\centering
\begin{tabular}{|l|c|}
\hline
\multicolumn{1}{|c|}{\textbf{Method}} & \textbf{RMSE (\$)} \\ \hline
Bessinger \& Jacobs \cite{bessinger2016quantifying} : HA & \textbf{29365.00} \\
Proposed Method : HA & 32108.91 $\pm$ 14.80 \\ \hline
Bessinger \& Jacobs \cite{bessinger2016quantifying} : IF & 53727.00 \\
Proposed Method : IF & \textbf{35188.72 $\pm$ 7.98} \\ \hline
Bessinger \& Jacobs \cite{bessinger2016quantifying} : HA + IF & 28281 \\
Proposed Method : HA + IF & \textbf{24439.64 $\pm$ 11.63} \\ \hline
\end{tabular}
\caption{Comparison of results for Housing price regression on the Fayette County house price data-set from~\cite{bessinger2016quantifying}. Superior performance is indicated by lower RMSE values. HA and IF stand for House Attributes and Image Features respectively.}
\vspace{-0.1in}
\label{table:ICIP16Comp}

\end{table}

\paragraph*{Effect of Feature Combinations:} We also experiment with using different feature combinations at the feature fusion stage of our approach. We leverage three classes of features: (i) deep features (DF) extracted from satellite images, (ii) house attributes (HA), and (iii) place of interest (POI) features. The results in terms of RMSE and $R^2$ are listed in Tables~\ref{table:featureCombRMSE} and~\ref{table:featureCombR2}. HA features are unable to capture price variations, as they characterise the house itself, but not its neighborhood. Estimators using DF or POI features, which capture the neighborhood information, are able to improve the prediction. When we combine HA with DF or POI features for estimation, a large improvement over the previous  configurations is seen. \textit{This observation indicates that house level attributes and neighborhood level features such as DF and POI are complementary in capturing price variations.}
An additional observation is that combining DF and POI features does not lead to any significant advance in  price estimation over the case where either of them are utilized, demonstrating that \textit{DF and POI features are highly correlated} in the context of price estimation. The results also highlight the difference between `sale' and `rent' sub-markets. For rental properties, HA by itself is able to explain the variation of prices, in contrast to the houses put up for sale. \textit{This indicates that transactions carried out for relatively short-term usage of houses place more value on the amenities of the house itself rather than its neighborhood.} 

\paragraph*{Effect of neighbourhood size:} We next conduct experiments on the effects of including information from progressively larger neighbourhoods of house samples through satellite images. The zoom values for Google Map queries are integer values ranging from 15 to 20. We present RMSE and $R^2$ results using the Random Forest estimator for the London data-set in Table~\ref{table:neighbourhoodSize}. \textit{The results show that the proposed method is able to better predict price by including larger neighbourhood contexts.} This is a significant observation, because most often, the spatial contiguity matrix specified in the SAR model only captures local neighborhood information, which is not sufficient to capture all the spatial relationships.

\paragraph*{Deep feature extraction across cities:} As described in Section \ref{sec:DCNNTraining}, the DCNNs were trained using data from London data-set due to limited number of samples in Birmingham and Liverpool data-sets. From Table \ref{table:SARvsPropRMSE&R2}, we can observe that the features extracted through networks trained on a specific data-set are effective in estimating housing prices in a different city. \textit{This implies that satellite image feature extraction is generic across cities in the same broad geographical region, the British Isles.} Further studies could examine how effective such feature extraction schemes would be for cities in more contrasting regions.

\begin{table*}[]
\footnotesize
\centering
\begin{tabular}{|c|c|c|c|c|}
\hline
\multirow{2}{*}{\textbf{Method}} & \multicolumn{2}{c|}{\textbf{London}} & \textbf{Birmingham} & \textbf{Liverpool} \\ \cline{2-5} 
 & Rent price & Sale price & Sale price & Sale price \\ \hline
\multicolumn{5}{|c|}{RMSE ($\pounds$)} \\ \hline
SAR (DT) & 159.18 & 282989.82 & 52739.86 & 58075.00 \\ \hline
SAR (K-NN, K = 10) & 159.85 & 284324.75 & 58494.00 & 53249.85 \\ \hline
Linear & 143.75 & 259411.88 & 38316.55 & 39395.46 \\ \hline
Random Forest & 80.04 $\pm$ 0.51 & 127328.66 $\pm$ 1204.43 & 27868.22 $\pm$ 968.52 & 31412.19 $\pm$ 565.78 \\ \hline
MLP & \textbf{74.52 $\pm$ 3.10} & \textbf{116639.78 $\pm$ 2941.28} & \textbf{20837.38 $\pm$ 140.60} & \textbf{29690.90 $\pm$ 408.67} \\ \hline
\multicolumn{5}{|c|}{$R^2$} \\ \hline
SAR (DT) & 0.5794 & 0.5596 & 0.5884 & 0.5231 \\ \hline
SAR (K-NN, K=10) & 0.5759 & 0.5554 & 0.5804 & 0.5957 \\ \hline
Linear & 0.6569 & 0.6298 & 0.7824 & 0.8165 \\ \hline
Random Forest & 0.8936 $\pm$ 0.0014 & 0.9108 $\pm$ 0.0017 & 0.8848 $\pm$ 0.0081 & 0.8833 $\pm$ 0.0042 \\ \hline
MLP & \textbf{0.9077 $\pm$ 0.0078} & \textbf{0.9251 $\pm$ 0.0038} & \textbf{0.9356 $\pm$ 0.0009} & \textbf{0.8958 $\pm$ 0.0029} \\ \hline
\end{tabular}
\caption{RMSE and $R^2$ for SAR and proposed housing price prediction models. Lower values for RMSE and Higher values for $R^2$ indicated superior performance with the method with best results highlighted in bold.}
\label{table:SARvsPropRMSE&R2}
\end{table*}

\begin{table*}[!t]
\footnotesize
\centering
\begin{tabular}{|c|c|c|c|c|}
\hline
\multirow{2}{*}{\textbf{\begin{tabular}[c]{@{}c@{}}Feature \\ combinations\end{tabular}}} & \multicolumn{2}{c|}{\textbf{London}} & \textbf{Birmingham} & \textbf{Liverpool} \\ \cline{2-5} 
 & \begin{tabular}[c]{@{}c@{}}Rent price \\ RMSE (\pounds)\end{tabular} & \begin{tabular}[c]{@{}c@{}}Sale price \\ RMSE (\pounds)\end{tabular} & \begin{tabular}[c]{@{}c@{}}Sale price \\ RMSE (\pounds)\end{tabular} & \begin{tabular}[c]{@{}c@{}}Sale price \\ RMSE (\pounds)\end{tabular} \\ \hline
\multicolumn{5}{|c|}{\textbf{Random Forest Regression}} \\ \hline
HA & 130.10 $\pm$ 0.49 & 366954.09 $\pm$ 130.14 & 60725.84 $\pm$ 165.94 & 72527.92 $\pm$ 123.17 \\ \hline
POI & 200.79 $\pm$ 0.06 & 202220.03 $\pm$ 1551.227 & 35647.28 $\pm$ 1290.66 & 38418.66 $\pm$ 429.54 \\ \hline
DF & 136.59 $\pm$ 0.63 & 213617.63 $\pm$ 1472.42 & 36169.24 $\pm$ 924.44 & 42262.78 $\pm$ 513.85 \\ \hline
HA + POI & \textbf{72.24 $\pm$ 0.69} & \textbf{108962.85 $\pm$ 769.53} & \textbf{19171.26 $\pm$ 841.28} & \textbf{25552.93 $\pm$ 406.72} \\ \hline
HA + DF & 80.04 $\pm$ 0.51 & 127328.66 $\pm$ 1204.43 & 27868.22 $\pm$ 968.52 & 31412.19 $\pm$ 565.78 \\ \hline
DF + POI & 135.72 $\pm$ 0.51 & 211056.00 $\pm$ 849.26 & 35588.73,+/- 547.50 & 38685.94 $\pm$ 485.08 \\ \hline
\multicolumn{5}{|c|}{\textbf{Multi-layer Perceptron Regression}} \\ \hline
HA & 204.48 $\pm$ 0.49 & 373608.36 $\pm$ 67.50 & 64004.49 $\pm$ 43.88 & 76499.74 $\pm$ 296.51 \\ \hline
POI & 183.93 $\pm$ 3.69 & 335122.36 $\pm$ 4552.90 & 70708.61 $\pm$ 466.11 & 69729.30 $\pm$ 656.34 \\ \hline
DF & 129.38 $\pm$ 2.74 & 190841.13 $\pm$ 1343.46 & 25132.21 $\pm$ 262.18 & 39789.38 $\pm$ 489.26 \\ \hline
HA + POI & 116.91 $\pm$ 3.68 & 213356.72 $\pm$ 5795.78 & 48862.52 $\pm$ 1022.05 & 53585.74 $\pm$ 598.36 \\ \hline
HA + DF & \textbf{74.52 $\pm$ 3.10} & \textbf{116639.78 $\pm$ 2941.28} & \textbf{20837.38 $\pm$ 140.60} & \textbf{29690.90 $\pm$ 408.67} \\ \hline
DF + POI & 128.07 $\pm$ 1.63 & 190945.77 $\pm$ 1443.09 & 25858.58 $\pm$ 215.43 & 38795.62 $\pm$ 310.24 \\ \hline
\end{tabular}
\caption{RMSE for different feature combinations with Random Forest and Multi-layer perceptron regression. Lower values indicated superior performance with the method with best results highlighted in bold. HA, POI and DF stand for House Attributes, Point of Interest and Deep Features respectively.}
\label{table:featureCombRMSE}
\end{table*}

\begin{table*}[!t]
\footnotesize
\centering
\begin{tabular}{|c|c|c|c|c|}
\hline
\multirow{2}{*}{\textbf{\begin{tabular}[c]{@{}c@{}}Feature \\ combinations\end{tabular}}} & \multicolumn{2}{c|}{\textbf{London}} & \textbf{Birmingham} & \textbf{Liverpool} \\ \cline{2-5} 
 & \begin{tabular}[c]{@{}c@{}}Rent price \\ $R^2$\end{tabular} & \begin{tabular}[c]{@{}c@{}}Sale price \\ $R^2$\end{tabular} & \begin{tabular}[c]{@{}c@{}}Sale price\\ $R^2$\end{tabular} & \begin{tabular}[c]{@{}c@{}}Sale price\\  $R^2$\end{tabular} \\ \hline
\multicolumn{5}{|c|}{\textbf{Random Forest Regression}} \\ \hline
HA & 0.7190 $\pm$ 0.0020 & 0.2594 $\pm$ 0.0005 & 0.4535 $\pm$ 0.0030 & 0.3781 $\pm$ 0.0021 \\ \hline
POI & 0.3307 $\pm$ 0.0004 & 0.7751 $\pm$ 0.0034 & 0.8114 $\pm$ 0.0136 & 0.8255 $\pm$ 0.0039 \\ \hline
DF & 0.6903 $\pm$ 0.0029 & 0.7490 $\pm$ 0.0035 & 0.8060 $\pm$ 0.0100 & 0.7888 $\pm$ 0.0051 \\ \hline
HA + POI & \textbf{0.9133 $\pm$ 0.0017} & \textbf{0.9347 $\pm$ 0.0009} & \textbf{0.9454 $\pm$ 0.0049} & \textbf{0.9228 $\pm$ 0.0024} \\ \hline
HA + DF & 0.8936 $\pm$ 0.0014 & 0.9108 $\pm$ 0.0017 & 0.8848 $\pm$ 0.0081 & 0.8833 $\pm$ 0.0042 \\ \hline
DF + POI & 0.6942 $\pm$ 0.0023 & 0.7550 $\pm$ 0.0020 & 0.8122, $\pm$ 0.0058 & 0.8230 $\pm$ 0.0044 \\ \hline
\multicolumn{5}{|c|}{\textbf{Multi-layer Perceptron Regression}} \\ \hline
\multicolumn{1}{|c|}{HA} & \multicolumn{1}{l|}{0.3059 $\pm$ 0.0033} & \multicolumn{1}{l|}{0.2323 $\pm$ 0.0003} & \multicolumn{1}{l|}{0.3929 $\pm$ 0.0008} & \multicolumn{1}{l|}{0.3081 $\pm$ 0.0054} \\ \hline
\multicolumn{1}{|c|}{POI} & \multicolumn{1}{l|}{0.4382 $\pm$ 0.0223} & \multicolumn{1}{l|}{0.3822 $\pm$ 0.0168} & \multicolumn{1}{l|}{0.2590 $\pm$,0.0098} & \multicolumn{1}{l|}{0.4251 $\pm$,0.0108} \\ \hline
\multicolumn{1}{|c|}{DF} & \multicolumn{1}{l|}{0.7220 $\pm$ 0.0119} & \multicolumn{1}{l|}{0.7997 $\pm$ 0.0028} & \multicolumn{1}{l|}{0.9064 $\pm$ 0.0020} & \multicolumn{1}{l|}{0.8128 $\pm$ 0.0046} \\ \hline
\multicolumn{1}{|c|}{HA + POI} & \multicolumn{1}{l|}{0.7729 $\pm$ 0.0142} & \multicolumn{1}{l|}{0.7495 $\pm$ 0.0137} & \multicolumn{1}{l|}{0.6460 $\pm$ 0.0147} & \multicolumn{1}{l|}{0.6605 $\pm$ 0.0076} \\ \hline
\multicolumn{1}{|c|}{HA + DF} & \multicolumn{1}{l|}{\textbf{0.9077 $\pm$ 0.0078}} & \multicolumn{1}{l|}{\textbf{0.9251 $\pm$ 0.0038}} & \multicolumn{1}{l|}{\textbf{0.9356 $\pm$ 0.0009}} & \multicolumn{1}{l|}{\textbf{0.8958 $\pm$ 0.0029}} \\ \hline
\multicolumn{1}{|c|}{DF + POI} & \multicolumn{1}{l|}{0.7277 $\pm$ 0.0069} & \multicolumn{1}{l|}{0.7995 $\pm$ 0.0030} & \multicolumn{1}{l|}{0.9009 $\pm$ 0.0017} & \multicolumn{1}{l|}{0.8220 $\pm$ 0.0028} \\ \hline
\end{tabular}
\caption{$R^2$ for different feature combinations with Random Forest and Multi-layer perceptron regression. Higher values indicated superior performance with the method with best results highlighted in bold. HA, POI and DF stand for House Attributes, Point of Interest and Deep Features respectively.}
\label{table:featureCombR2}
\end{table*}

\section{Discussion and Conclusion}\label{section:discussions}

From the experiments, we can see that joint representations of houses in terms of their individual attributes 
(HA) and the neighbourhood around them improves the accuracy of housing price prediction. This result is reinforced by conclusions from earlier works such as \cite{chopra2007discovering}. PoI and DF features have been consistently shown to be positively correlated in their effects on house prices across estimators and city data-sets. Hence, multiscale DCNN-derived features from publicly available satellite imagery could be used in place of PoI data, which are usually of proprietary nature and require explicit annotations regarding local businesses.

As we can see from the results, utilizing deep features from satellite images at different geo-spatial resolutions leads to comparable or superior performance compared to using either latitude-longitude information explicitly in SAR models or utilizing Point of interest features. Also, employing information from images of larger areas surrounding house samples leads to improved accuracy in housing price estimation.

\section{Acknowledgements} 
Research was sponsored by the Army Research Laboratory and was accomplished under Cooperative Agreement Number W911NF-09-2-0053 (the ARL Network Science CTA). The views and conclusions contained in this document are those of the authors and should not be interpreted as representing the official policies, either expressed or implied, of the Army Research Laboratory or the U.S. Government. The U.S. Government is authorized to reproduce and distribute reprints for Government purposes notwithstanding any copyright notation here on.

{\small
\bibliographystyle{ieee}
\bibliography{egbib}

\begin{thebibliography}{10}\itemsep=-1pt

\bibitem{abadi2016tensorflow}
M.~Abadi, A.~Agarwal, P.~Barham, E.~Brevdo, Z.~Chen, C.~Citro, G.~S. Corrado,
  A.~Davis, J.~Dean, M.~Devin, et~al.
\newblock Tensorflow: Large-scale machine learning on heterogeneous distributed
  systems.
\newblock {\em arXiv preprint arXiv:1603.04467}, 2016.

\bibitem{arietta2014city}
S.~M. Arietta, A.~A. Efros, R.~Ramamoorthi, and M.~Agrawala.
\newblock City forensics: Using visual elements to predict non-visual city
  attributes.
\newblock {\em IEEE transactions on visualization and computer graphics},
  20(12):2624--2633, 2014.

\bibitem{bessinger2016quantifying}
Z.~Bessinger and N.~Jacobs.
\newblock Quantifying curb appeal.
\newblock In {\em 2016 IEEE International Conference on Image Processing
  (ICIP)}, pages 4388--4392. IEEE, 2016.

\bibitem{bourassa2007spatial}
S.~C. Bourassa, E.~Cantoni, and M.~Hoesli.
\newblock Spatial dependence, housing submarkets, and house price prediction.
\newblock {\em The Journal of Real Estate Finance and Economics},
  35(2):143--160, 2007.

\bibitem{chen2012four}
Y.~Chen.
\newblock On the four types of weight functions for spatial contiguity matrix.
\newblock {\em Letters in Spatial and Resource Sciences}, 5(2):65--72, 2012.

\bibitem{cheng2016survey}
G.~Cheng and J.~Han.
\newblock A survey on object detection in optical remote sensing images.
\newblock {\em ISPRS Journal of Photogrammetry and Remote Sensing}, 117:11--28,
  2016.

\bibitem{chopra2007discovering}
S.~Chopra, T.~Thampy, J.~Leahy, A.~Caplin, and Y.~LeCun.
\newblock Discovering the hidden structure of house prices with a
  non-parametric latent manifold model.
\newblock In {\em Proceedings of the 13th ACM SIGKDD international conference
  on Knowledge discovery and data mining}, pages 173--182. ACM, 2007.

\bibitem{GoogleMaps}
{Google Inc.}
\newblock {\em Google Maps API}, 2016 (accessed October 6th, 2016).
\newblock https://developers.google.com/maps/.

\bibitem{GooglePlaces}
{Google Inc.}
\newblock {\em Google Places API}, 2016 (accessed October 6th, 2016).
\newblock https://developers.google.com/maps/.

\bibitem{hu2007road}
J.~Hu, A.~Razdan, J.~C. Femiani, M.~Cui, and P.~Wonka.
\newblock Road network extraction and intersection detection from aerial images
  by tracking road footprints.
\newblock {\em IEEE Transactions on Geoscience and Remote Sensing},
  45(12):4144--4157, 2007.

\bibitem{jean2016poverty}
N.~Jean, M.~Burke, M.~Xie, W.~M. Davis, D.~B. Lobell, and S.~Ermon.
\newblock {Combining satellite imagery and machine learning to predict
  poverty}.
\newblock {\em Science}, 353:790--794, 2016.

\bibitem{fineTuning}
S.~Karayev, M.~Trentacoste, H.~Han, A.~Agarwala, T.~Darrell, A.~Hertzmann, and
  H.~Winnemoeller.
\newblock Recognizing image style.
\newblock In {\em Proceedings of the British Machine Vision Conference}. BMVA
  Press, 2014.

\bibitem{CVPR14_Khosla}
A.~Khosla, B.~An, J.~J. Lim, and A.~Torralba.
\newblock Looking beyond the visible scene.
\newblock In {\em IEEE Conference on Computer Vision and Pattern Recognition
  (CVPR)}, Ohio, USA, June 2014.

\bibitem{kibriya2007empirical}
A.~M. Kibriya and E.~Frank.
\newblock An empirical comparison of exact nearest neighbour algorithms.
\newblock In {\em European Conference on Principles of Data Mining and
  Knowledge Discovery}, pages 140--151. Springer, 2007.

\bibitem{kockelman1997effects}
K.~Kockelman.
\newblock Effects of location elements on home purchase prices and rents in san
  francisco bay area.
\newblock {\em Transportation Research Record: Journal of the Transportation
  Research Board}, (1606):40--50, 1997.

\bibitem{krizhevsky2012imagenet}
A.~Krizhevsky, I.~Sutskever, and G.~E. Hinton.
\newblock Imagenet classification with deep convolutional neural networks.
\newblock In {\em Advances in neural information processing systems}, pages
  1097--1105, 2012.

\bibitem{krupka2009neighborhood}
D.~J. Krupka and D.~Noonan.
\newblock Neighborhood dynamics and the housing price effects of spatially
  targeted economic development policy.
\newblock {\em SSRN Working Paper Series}, 2009.

\bibitem{lesage2009introduction}
J.~P. LeSage and R.~K. Pace.
\newblock {\em Introduction to Spatial Econometrics (Statistics, textbooks and
  monographs)}.
\newblock CRC Press, 2009.

\bibitem{meng2012object}
L.~Meng and J.~P. Kerekes.
\newblock Object tracking using high resolution satellite imagery.
\newblock {\em IEEE Journal of Selected Topics in Applied Earth Observations
  and Remote Sensing}, 5(1):146--152, 2012.

\bibitem{BingMaps}
{Microsoft Corporation}.
\newblock {\em Bing Maps: Developer Resources}, 2016 (accessed October 6th,
  2016).
\newblock https://www.microsoft.com/maps/developer-resources.aspx.

\bibitem{mnih2010learning}
V.~Mnih and G.~E. Hinton.
\newblock Learning to detect roads in high-resolution aerial images.
\newblock In {\em European Conference on Computer Vision}, pages 210--223.
  Springer, 2010.

\bibitem{ordonez2014learning}
V.~Ordonez and T.~L. Berg.
\newblock Learning high-level judgments of urban perception.
\newblock In {\em European Conference on Computer Vision}, pages 494--510.
  Springer, 2014.

\bibitem{osland2010application}
L.~Osland.
\newblock An application of spatial econometrics in relation to hedonic house
  price modeling.
\newblock {\em Journal of Real Estate Research}, 2010.

\bibitem{scikit-learn}
F.~Pedregosa, G.~Varoquaux, A.~Gramfort, V.~Michel, B.~Thirion, O.~Grisel,
  M.~Blondel, P.~Prettenhofer, R.~Weiss, V.~Dubourg, J.~Vanderplas, A.~Passos,
  D.~Cournapeau, M.~Brucher, M.~Perrot, and E.~Duchesnay.
\newblock Scikit-learn: Machine learning in {P}ython.
\newblock {\em Journal of Machine Learning Research}, 12:2825--2830, 2011.

\bibitem{Redfin}
{Redfin}.
\newblock Redfin: Real estate, homes for sale, mls listings, agents, 2016
  (accessed October 6th, 2016).
\newblock https://www.redfin.com/.

\bibitem{ILSVRC15}
O.~Russakovsky, J.~Deng, H.~Su, J.~Krause, S.~Satheesh, S.~Ma, Z.~Huang,
  A.~Karpathy, A.~Khosla, M.~Bernstein, A.~C. Berg, and L.~Fei-Fei.
\newblock {ImageNet Large Scale Visual Recognition Challenge}.
\newblock {\em International Journal of Computer Vision (IJCV)},
  115(3):211--252, 2015.

\bibitem{simonyan2014very}
K.~Simonyan and A.~Zisserman.
\newblock Very deep convolutional networks for large-scale image recognition.
\newblock In {\em International Conference on Learning Representations}, 2015.

\bibitem{steenbeek2011longitudinal}
W.~Steenbeek and J.~R. Hipp.
\newblock A longitudinal test of social disorganization theory: Feedback
  effects among cohesion, social control, and disorder.
\newblock {\em Criminology}, 49(3):833--871, 2011.

\bibitem{szegedy2015rethinking}
C.~Szegedy, V.~Vanhoucke, S.~Ioffe, J.~Shlens, and Z.~Wojna.
\newblock Rethinking the inception architecture for computer vision.
\newblock {\em arXiv preprint arXiv:1512.00567}, 2015.

\bibitem{Trulia}
{Trulia}.
\newblock Trulia: Real estate listings, homes for sale, housing data, 2016
  (accessed October 6th, 2016).
\newblock https://www.trulia.com/.

\bibitem{Zillow}
{Zillow Group}.
\newblock Zillow: Real estate, apartments, mortgages \& home values, 2016
  (accessed October 6th, 2016).
\newblock http://www.zillow.com/.

\bibitem{Zoopla}
{Zoopla PLC}.
\newblock {\em Zoopla: Search Property to Buy, Rent, House Prices, Estate
  Agents}, 2016 (accessed October 6th, 2016).
\newblock http://www.zoopla.co.uk.

\bibitem{zwack2011modeling}
L.~M. Zwack, C.~J. Paciorek, J.~D. Spengler, and J.~I. Levy.
\newblock Modeling spatial patterns of traffic-related air pollutants in
  complex urban terrain.
\newblock {\em Environmental Health Perspectives}, 119(6):852, 2011.

\end{thebibliography}
}

\end{document}